\documentclass[table]{article}

\usepackage{arxiv}

\usepackage{amsmath}

\usepackage[utf8]{inputenc} 
\usepackage[T1]{fontenc}    
\usepackage{hyperref}       
\usepackage{url}            
\usepackage{booktabs}       
\usepackage{amsfonts}       
\usepackage{nicefrac}       
\usepackage{microtype}      
\usepackage{cleveref}       
\usepackage{graphicx}
\usepackage[numbers, square, sort&compress]{natbib} 
\usepackage{doi} 

\usepackage{tikz}
\usetikzlibrary{fadings}
\usepackage{tcolorbox}
\usepackage{mathtools}
\usepackage{amsthm}
\usepackage{bbm}
\usepackage{bm}
\usepackage{tabularx}
\usepackage{threeparttable}
\usepackage{makecell}
\usepackage{multirow}
\usepackage{booktabs}
\usepackage{subcaption}
\usepackage{flushend}
\usepackage{dirtytalk}
\usepackage{xcolor}
\usepackage{circuitikz}
\usepackage{amsmath}
\usepackage{algorithm}
\usepackage{algorithmic}
\usepackage{enumitem}
\usepackage{hyperref}

\usetikzlibrary{arrows.meta,positioning,fit,shapes.geometric,calc}

\definecolor{ubblue}{HTML}{004E9F}   
\definecolor{ubyellow}{HTML}{FCBA00} 
\definecolor{ubgrey}{HTML}{909085}   

\title{Automated Regulatory Compliance Question Answering in Financial Services with Domain-Adapted Retrieval-Augmented Generation}

\date{}

\usepackage{authblk}

\author[1,2]{%
	Tobias Deu{\ss}er\thanks{\texttt{tdeusser@uni-bonn.de}, ORCID-ID: 0000-0003-4685-0847}%
}
\author[1]{%
	Abhishek Pillai
}
\author[3]{%
	Aurelio F. Bariviera
}
\author[3]{%
	Dhananjay Bhardwaj
}
\author[1,2]{%
	\\Lorenz Sparrenberg
}
\author[2,4]{%
	David Berghaus
}
\author[1,2,4]{%
	Christian Bauckhage
}
\author[1,2,4]{%
	Rafet Sifa
}

\affil[1]{University of Bonn, Bonn, Germany}
\affil[2]{Lamarr-Institute for Machine Learning and Artificial Intelligence, Bonn, Germany}
\affil[3]{Universitat Rovira i Virgili, Reus, Spain}
\affil[4]{Fraunhofer IAIS, Sankt Augustin, Germany}

\renewcommand{\headeright}{}
\renewcommand{\undertitle}{}
\renewcommand{\shorttitle}{Automated Regulatory Compliance Question Answering in Financial Services with Domain-Adapted RAG}

\hypersetup{
pdftitle={Automated Regulatory Compliance Question Answering in Financial Services with Domain-Adapted Retrieval-Augmented Generation},
pdfsubject={cs.CL, cs.AI},
pdfauthor={Tobias Deu{\ss}er, Abhishek Pillai, Aurelio F. Bariviera, Dhananjay Bhardwaj, Lorenz Sparrenberg, David Berghaus, Christian Bauckhage, Rafet Sifa},
pdfkeywords={retrieval-augmented generation, regulatory compliance, legal NLP, hybrid retrieval, natural language processing},
}

\begin{document}
\maketitle
\begin{abstract}
Financial institutions operate under dense, frequently amended rulebooks, and answering a compliance question correctly requires not only fluency but verifiable grounding in the authoritative text. Large language models are attractive for this task, yet the models that firms can realistically deploy on-premise are compact ones, and compact models hallucinate obligations. We study whether a carefully domain-adapted retrieval-augmented generation pipeline closes that gap. Our retriever is built in three stages on top of LegalBERT: entailment tuning that recasts question--passage matching as premise--hypothesis reconstruction, contrastive tuning with in-batch negatives, and score-level fusion with BM25. Our generator is a compact model (2B--12B parameters) served under 4-bit quantization, either prompted or adapted with retrieval-aware fine-tuning (RAFT) through LoRA. On ObliQA, a question-answering benchmark built from the Abu Dhabi Global Market rulebooks, the staged retriever raises Recall@10 from 0.256 to 0.774 and outperforms BM25 (0.678) and E5-large-v2 (0.758), the strongest general-purpose dense encoder we tested. 
RAFT-LoRA then improves the composite RePASs answer-quality score for every model we could adapt, with the largest gain on the weakest one. However, the adapted models do not transfer to Australian case-law questions, and a closed-book model that receives no passages at all scores within 0.011 RePASs of the full pipeline while producing answers that cite nothing and misstate obligations. The retrieval gain is therefore measured directly, the generation gain is a gain in RePASs rather than demonstrated grounding, and grounding itself requires an evaluation protocol that RePASs does not provide.
\end{abstract}


\section{Introduction}

Regulated financial companies operate inside a body of rules that is large, cross-referential, and in constant flux \cite{jeong2026dataset}. A single question is typically answered by locating a handful of facts scattered across chapters and reconciling them. These financial companies currently do this through specialist staff and commercial research tools, which makes compliance slow and, when a provision is missed, expensive \cite{sifa2019towards,von2021integrating,raj2021alibert}.

Large language models (LLMs) are an obvious candidate for automating parts of this workflow \cite{berger2023towards,deusser2023uncovering,naveed2024comprehensiveoverviewlargelanguage,deusser2025leveraging}. Used out of the box, however, they are a poor fit. Without access to the source text, an LLM has to rely on what it memorized during training, and its recall of specific rulebook clauses is unreliable: it may paraphrase provisions that do not exist or omit binding conditions, and it sounds equally confident in both cases~\cite{bommasani2022opportunitiesrisksfoundationmodels}. In compliance work this is particularly problematic, because such an answer cannot be traced to any provision and therefore cannot be checked.

Retrieval-augmented generation (RAG)~\cite{lewis2021retrievalaugmentedgenerationknowledgeintensivenlp} addresses this problem by retrieving relevant passages from an authoritative corpus and supplying them as context, so that answers can be traced back to the source text. The quality of a RAG system, however, depends heavily on its retriever. Provisions in different chapters of a rulebook often use nearly identical wording while differing in scope, applicability, or exemptions. Lexical overlap is therefore a weak relevance signal, and general-purpose dense encoders trained on web data are not designed to tell which of two similar clauses applies to a given case~\cite{thakur2021beirheterogenousbenchmarkzeroshot}.

A second constraint is a more practical one. Regulatory documents and the questions asked about them are often confidential, so many institutions cannot send queries to a hosted frontier model and have to run models on their own hardware~\cite{deusser2025resource}. In practice, this limits the choice to quantized models with up to roughly 12B parameters (depending on the budget of the firm of course), which have the least legal knowledge and the weakest grounding behavior. Improvements therefore have to come from the pipeline around the model rather than from model scale.

In this paper, we investigate how far such a pipeline can be improved. We build a RAG system for regulatory question answering in which both halves, the retrieval and the generation, are domain-adapted, and we evaluate it on ObliQA \cite{gokhan2024riragregulatoryinformationretrieval}, a benchmark derived from the rulebooks of the Abu Dhabi Global Market (ADGM) and on Open Australian LegalQA \cite{butler-2025-open-australian-legal-corpus} as an out-of-domain control. Our contributions are:

\begin{itemize}
\item \textbf{A three-stage retriever} that starts from LegalBERT \cite{chalkidis2020legalbertmuppetsstraightlaw} and applies entailment tuning, contrastive tuning, and linear score fusion with BM25. Each stage helps, and the composition lifts Recall@10 on ObliQA from 0.256 to 0.774, past BM25 (0.678) and past E5-large-v2 (0.758) despite using a far smaller encoder.

\item \textbf{A leakage-free re-split} of ObliQA. The published splits share source documents across train, validation, and test; we rebuild them at document level and quantify what this costs in dataset size.

\item \textbf{A controlled comparison of prompting against retrieval-aware fine-tuning} for five compact generators. Prompting effects are model-dependent and often negative, whereas RAFT-LoRA improves the composite RePASs score for all three models we could adapt stably, most for the weakest one.

\item \textbf{Two negative results.} Adapters trained on ADGM rulebook data do not transfer to Australian case law, and a closed-book model without access to any passage scores within 0.011 RePASs of the full retrieval pipeline, even though its answers are not grounded. We attribute the latter to a weakness of the metric and discuss what this means for evaluating regulatory QA systems.

\end{itemize}

\section{Related Work}

\subsection{Regulatory and Legal Question Answering}

Regulatory NLP applies language technology to statutes, rulebooks, and compliance guidance, where texts are long, heavily structured, and full of normative operators (obligations, permissions, prohibitions) whose scope is set by conditions and exceptions \cite{zhong2020doesnlpbenefitlegal}. Answering questions over them is closer to multi-passage aggregation than to span extraction, since the governing conditions for one obligation are routinely stated elsewhere.

Progress in this area has been driven by datasets. CUAD \cite{hendrycks2021cuadexpertannotatednlpdataset} annotates contract clauses for review tasks; LexGLUE \cite{chalkidis2021lexglue} assembles legal understanding benchmarks; FinQA \cite{chen2022finqadatasetnumericalreasoning} targets numerical reasoning over financial filings. Closest to our setting is ObliQA \cite{gokhan2024riragregulatoryinformationretrieval}, which pairs questions with the ADGM provisions that answer them and introduces RePASs, an obligation-aware answer metric; that work also defines the retrieve-then-generate task formulation we adopt. \cite{chalkidis2021regulatorycompliancedoc2docinformation} treat regulatory compliance as document-to-document retrieval over European legislation, and \cite{abdullin-etal-2023-synthetic} study synthetic data for the same domain. \cite{bashir2026domainadaptationsyntheticdatafinetuning} generate question--answer pairs from German statutes and adapt LLMs with parameter-efficient fine-tuning, reporting gains over unadapted baselines. That strategy is close in spirit to the fine-tuning half of our pipeline, though it is applied to a different jurisdiction and without a retrieval-aware objective. Earlier legal QA systems relied on rules and keyword matching \cite{sergot1986british,turtle1995text,quaresma2005question}, later on neural judgment and retrieval models \cite{chalkidis2019neurallegaljudgmentprediction,zhong2019jecqalegaldomainquestionanswering,zheng2021doespretraininghelpassessing}. A broader and more detailed survey of LLM use in finance is given in \cite{choi2025large}.

\subsection{Sparse, Dense, and Hybrid Retrieval}

BM25 \cite{robertson-1994-okapi} remains a strong baseline wherever terminology is fixed, which describes regulatory text well, but it cannot bridge paraphrase. Dense retrieval \cite{karpukhin2020densepassageretrievalopendomain} embeds queries and passages into a shared space and handles paraphrase, and general-purpose encoders such as E5 \cite{wang2024multilinguale5textembeddings} and BGE \cite{luo2024bgelandmarkembeddingchunkingfree} transfer well across many domains. Specifically domain-pretrained encoders such as LegalBERT \cite{chalkidis2020legalbertmuppetsstraightlaw} know legal vocabulary but, as we confirm in Section~\ref{sec:retrieval-results}, are not good retrievers out of the box: masked-language pretraining optimizes token prediction, not the geometry of a similarity space.

Two lines of work address this limitation. Contrastive training with in-batch negatives and a multiple-negatives ranking loss \cite{reimers2019sentencebertsentenceembeddingsusing,henderson2017efficientnaturallanguageresponse,chen2020simpleframeworkcontrastivelearning} can reshape the embedding space directly. Entailment tuning \cite{dai2024improvedensepassageretrieval} instead reformulates retrieval as an inference task: a passage is relevant if it entails a claim derived from the question, which is a better match for legal text where two passages can be topically identical yet only one legally supports the answer. Combining sparse and dense scores recovers the strengths of both \cite{luo2023studyefficiencygeneralizationlight,wang2021interpolation}. Our retriever composes all three ideas in sequence rather than choosing among them.

\subsection{Retrieval-Augmented Generation}

RAG \cite{lewis2021retrievalaugmentedgenerationknowledgeintensivenlp,gao2024retrievalaugmentedgenerationlargelanguage} conditions generation on retrieved evidence, reducing hallucination and making answers attributable \cite{izacard2021leveragingpassageretrievalgenerative,hillebrand2024advancing}. Extensions target its failure modes: Self-RAG \cite{asai2023selfraglearningretrievegenerate} adds self-critique, Robust-RAG \cite{xiang2024certifiablyrobustragretrieval} defends against corrupted contexts, and RQ-RAG \cite{chan2024rqraglearningrefinequeries} refines ambiguous queries. RAFT \cite{zhang2024raftadaptinglanguagemodel} takes a different route and trains the generator itself to read a retrieved context: each training instance mixes an oracle passage with distractors, and for a fraction of instances withholds the oracle entirely, so the model learns to use evidence when present and to abstain from inventing it when absent. We adopt RAFT because it specifically targets hallucinations caused by irrelevant retrieved passages, which a high-recall retriever over regulatory text will inevitably return.

\subsection{Parameter-Efficient Fine-Tuning}
Full fine-tuning of a 7B model requires considerably more GPU memory than many compliance teams likely have available. Adapter modules \cite{houlsby2019parameterefficienttransferlearningnlp} and LoRA \cite{hu2021loralowrankadaptationlarge}, which learns a low-rank update to frozen attention projections, reduce the trainable parameter count by orders of magnitude, and QLoRA \cite{dettmers2023qloraefficientfinetuningquantized} adds 4-bit quantization of the frozen base so that a 7B model can be adapted on a single commodity GPU. Refinements continue in this direction \cite{qin2024accuratelorafinetuningquantizationllms,guo2024lqloralowrankplusquantized}. We use QLoRA throughout, which allows us to run all RAFT experiments on a single GPU and matches the on-premise setting described in the Introduction section.

\subsection{Positioning of this work}

Each component of our pipeline is established: domain-pretrained legal encoders \cite{chalkidis2020legalbertmuppetsstraightlaw}, entailment tuning \cite{dai2024improvedensepassageretrieval}, contrastive tuning with in-batch negatives \cite{reimers2019sentencebertsentenceembeddingsusing}, sparse--dense fusion \cite{wang2021interpolation}, and retrieval-aware fine-tuning \cite{zhang2024raftadaptinglanguagemodel}. We claim none of them as new. 

What is new is their composition into a single pipeline for financial-regulatory QA under an on-premise compute budget, together with a controlled evaluation of that pipeline: a staged retriever ablation on a leakage-free re-split, a like-for-like comparison of prompting against RAFT-LoRA across five compact generators, a transfer test of the resulting adapters to a different legal genre, and a closed-book control that isolates how much of the measured answer quality is attributable to retrieval at all. LLeQA \cite{louis2024interpretable} fixes a retrieve-then-read setup and varies the reader. CBR-RAG \cite{wiratunga2024cbr} varies the retrieval representation for a fixed generator. CLERC \cite{hou-etal-2025-clerc} reports retrieval and generation separately but with off-the-shelf retrievers. LegalBench-RAG \cite{pipitone2024legalbenchragbenchmarkretrievalaugmentedgeneration} isolates retrieval rigorously while stopping short of generation. We instead vary both halves over a single shared retrieval run, and add a closed-book control, so that the contribution of each stage, and the point at which the metric stops tracking grounding, are separately identifiable.

\section{Methodology}

\subsection{Task and Pipeline}

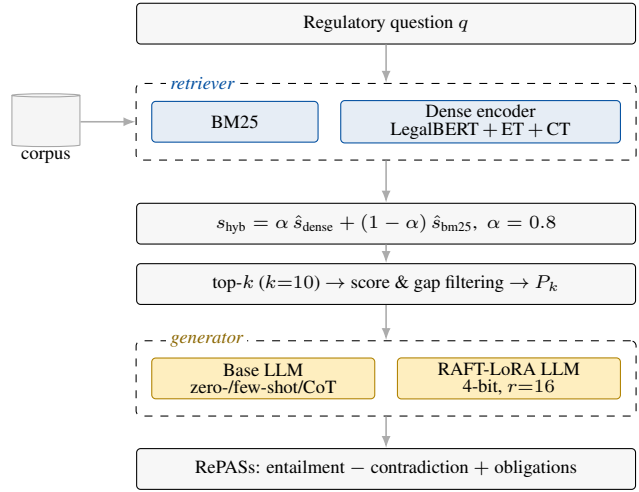
\begin{figure}[t]
\centering
\begin{tikzpicture}[
  font=\scriptsize,
  box/.style={draw, rounded corners=1.5pt, align=center,
              minimum height=5.4mm, inner xsep=3pt, inner ysep=1.6pt},
  wide/.style={box, minimum width=66mm, fill=gray!6},
  ret/.style={box, fill=ubblue!10, draw=ubblue},
  gen/.style={box, fill=ubyellow!25, draw=ubyellow!70!black},
  grp/.style={draw, dashed, rounded corners=2pt, inner xsep=2mm, inner ysep=2.2mm},
  tag/.style={fill=white, inner xsep=1.5pt, inner ysep=0.5pt, font=\scriptsize\itshape},
  ar/.style={-{Latex[length=1.6mm]}, gray!65, semithick},
]
\def\ax{6mm}

\node[wide] (q) at (\ax,0) {Regulatory question $q$};

\node[ret, minimum width=22mm] (bm) at (\ax-20mm,-13mm) {BM25};
\node[ret, minimum width=37mm] (dn) at (\ax+12.5mm,-13mm)
      {Dense encoder\\[-1.5pt] LegalBERT\,$+$\,ET\,$+$\,CT};
\node[grp, fit=(bm)(dn)] (retbox) {};
\node[tag, text=ubblue, anchor=west] at ([xshift=4mm]retbox.north west) {retriever};

\node[cylinder, draw=gray!70, shape border rotate=90, aspect=0.18,
      minimum width=9mm, minimum height=7mm, fill=gray!8,
      label={[font=\scriptsize, inner sep=1pt]below:corpus}] (corp)
      at (\ax-45mm,-13mm) {};

\node[wide] (fuse) at (\ax,-26.5mm)
      {$s_{\text{hyb}}=\alpha\,\hat{s}_{\text{dense}}+(1-\alpha)\,\hat{s}_{\text{bm25}}$,\ \ $\alpha=0.8$};
\node[wide] (filt) at (\ax,-34.5mm)
      {top-$k$ ($k{=}10$) $\rightarrow$ score \& gap filtering $\rightarrow$ $P_k$};

\node[gen, minimum width=29.5mm] (base) at (\ax-16.25mm,-47mm)
      {Base LLM\\[-1.5pt] zero-/few-shot/CoT};
\node[gen, minimum width=29.5mm] (raft) at (\ax+16.25mm,-47mm)
      {RAFT-LoRA LLM\\[-1.5pt] 4-bit, $r{=}16$};
\node[grp, fit=(base)(raft)] (genbox) {};
\node[tag, text=ubyellow!55!black, anchor=west] at ([xshift=4mm]genbox.north west) {generator};

\node[wide] (ev) at (\ax,-59mm)
      {RePASs: entailment $-$ contradiction $+$ obligations};

\draw[ar] (q.south)       -- (retbox.north);
\draw[ar] (retbox.south)  -- (fuse.north);
\draw[ar] (fuse.south)    -- (filt.north);
\draw[ar] (filt.south)    -- (genbox.north);
\draw[ar] (genbox.south)  -- (ev.north);
\draw[ar] (corp.east)     -- (corp.east -| retbox.west);
\end{tikzpicture}
\caption{The regulatory QA pipeline. Retrieval and generation are separate modules over a shared top-$k$ context, so a single retrieval run can be reused across every generation strategy.}
\label{fig:pipeline}
\end{figure}

Let $\mathcal{P}$ be a corpus of regulatory passages and $q$ a natural-language compliance question. The system must return an answer $a$ that is entailed by $\mathcal{P}$ and that covers the obligations $\mathcal{P}$ imposes on the situation in $q$. We factor this into a retriever $R: q \mapsto P_k \subset \mathcal{P}$ and a generator $L: (q, P_k) \mapsto a$, trained and evaluated separately so that the contribution of each is identifiable. Figure~\ref{fig:pipeline} shows the arrangement.

\subsection{Leakage-Free Data Construction}
\label{sec:splits}

A precondition for every result below is that no passage seen at evaluation time was used in training. ObliQA does not satisfy this. It ships with predefined splits, but the sets of source documents behind them intersect: $|D_{\text{train}} \cap D_{\text{eval}}|>0$ for every pair. Because ObliQA questions are generated from passages, a model that has seen one question from a document has effectively seen the passage that answers a different question from the same document. This is passage-level train--test leakage, and it inflates retrieval scores in particular, since the gold passage is then an item the encoder was optimised on rather than an unseen target. We therefore re-split at document level, which excludes this form of leakage by construction.

We therefore pool all questions and re-split at document level. The 40 ADGM documents are shuffled under a fixed seed and assigned 80\%, 15\%, and 5\% to train, validation, and test, respectively. Single-passage questions follow their document. Multi-passage questions are admitted only if all of their passages land in the same split, and are otherwise discarded. This removes 989 question--passage pairs, which we consider an acceptable trade-off for a leakage-free evaluation. All numbers in this paper are computed on these re-split sets and are therefore not directly comparable to published ObliQA results measured on the original splits. Table~\ref{tab:datasets} shows the resulting splits. Corpus construction is independent of the split: all 40 documents contribute their 13,705 passages, keyed by \texttt{documentID-passageID}.

We then audited the new splits for residual overlap with exact matching, fuzzy matching at 85\%, and embedding cosine similarity at 0.85. The first two find nothing. The third flags 139 train--eval question pairs, but inspection shows these are distinct obligations phrased alike (liquidity-risk stress testing appears under several rules), which is a property of regulatory drafting rather than leakage, so we retain them.

Open Australian LegalQA is converted to the same schema. After removing 2 duplicate questions from 2,124 records, we group by legal citation (e.g.\ \emph{Nasr v NRMA Insurance [2006] NSWSC 1018}) as the document identifier and split 80/20 by group, giving 1,695 training and 427 evaluation items over a 2,114-passage corpus. The same overlap audit finds 13 fuzzy and 16 semantic near-duplicates, all of which are distinct questions about parallel instruments, such as consecutive tariff concession orders or successive airworthiness directives, and are likewise kept.

\begin{table}[t]
\centering
\caption{Datasets after leakage-free re-splitting. ObliQA splits are disjoint at document level; Open Australian LegalQA is grouped by legal citation.}
\label{tab:datasets}
\footnotesize
\setlength{\tabcolsep}{4pt}
\begin{tabular}{l l r r r}
    \toprule
    Dataset & Split & Documents & Questions & Corpus passages \\
    \midrule

    ObliQA (ADGM) & train & 32 & 20{,}573 & \multirow{3}{*}{13{,}705} \\
    \rowcolor{gray!10}& validation & 6 & 3{,}191 & \\
    & test & 2 & 3{,}116 & \\

    \midrule

    \rowcolor{gray!10}Open Australian & train & -- & 1{,}695 & \multirow{2}{*}{2{,}114} \\
    LegalQA & evaluation & -- & 427 & \\

    \bottomrule
\end{tabular}

\end{table}

\subsection{Stage 1: Entailment Tuning}
\label{subsec:entailment_tuning}

The base encoder is LegalBERT \cite{chalkidis2020legalbertmuppetsstraightlaw}. Its pretraining objective gives it legal vocabulary but no notion of query--passage geometry, so we first retrain it on a task that is structurally closer to retrieval in this domain: deciding whether a passage \emph{supports} a claim.

Each question $q$ is rewritten into a declarative claim $c=f(q)$ by a rule-based mapping over six question types (\emph{when}, \emph{why}, \emph{who}, \emph{where}, \emph{does}, \emph{how}) with a generic fallback, so that a \emph{who} question becomes an assertion about the responsible entity. The gold passage $p^{+}$ becomes the premise and $c$ the hypothesis, giving every training instance the natural-language inference form
\begin{equation}
X = \texttt{"}\,\langle p^{+} \rangle\ \text{entails that}\ \langle H_{\text{masked}} \rangle\,\texttt{"} .
\end{equation}
Each hypothesis token is independently replaced by \texttt{[MASK]} with probability $\beta=0.8$, far above conventional MLM rates, and the model is trained to reconstruct only the masked positions:
\begin{equation}
\mathcal{L}_{\text{mlm}} = -\!\!\sum_{i \in \{\text{MASK}\}}\!\! \log P(\hat{h}_i = h_i \mid X).
\end{equation}
The high masking rate is intentional. When most of the hypothesis is masked, the model cannot simply copy tokens and has to use information from the premise to reconstruct it, which is the kind of query--passage matching that retrieval requires~\cite{dai2024improvedensepassageretrieval}. Training runs for 5 epochs (4 on Open Australian LegalQA) with AdamW at $2\times10^{-5}$, mixed precision, and gradient clipping at 1.0. The tuned encoder is converted into a bi-encoder by mean-pooling token embeddings under the SentenceTransformer interface.

\subsection{Stage 2: Contrastive Tuning}

Entailment tuning teaches the model whether a passage supports a claim, but it does not explicitly separate relevant from irrelevant passages in the embedding space. Stage 2 optimises the embedding space directly with the multiple-negatives ranking loss over in-batch negatives \cite{reimers2019sentencebertsentenceembeddingsusing}. For a batch of $N$ query--passage pairs,
\begin{equation}
\mathcal{L} = -\log \frac{\exp(\mathrm{sim}(q_i, p_i^{+})/\tau)}{\sum_{j=1}^{N} \exp(\mathrm{sim}(q_i, p_j)/\tau)},
\label{eq:mnrl}
\end{equation}
with $\mathrm{sim}$ the cosine similarity and $\tau$ a temperature. Queries and passages carry the prefixes \texttt{query:} and \texttt{passage:}; the same prefixes are used at inference, since mismatched prefixes degrade retrieval performance. Batch size is 16, learning rate $2\times10^{-5}$ with 100 warm-up steps, 2 epochs on ObliQA and 3 on Open Australian LegalQA, selected on validation Recall@10 from schedules up to 5 epochs.

We deliberately do not mine hard negatives, although we implemented mining with E5-large-v2 and a BGE reranker. We made this choice for two reasons. First, ObliQA is a multi-passage dataset: a passage that is semantically adjacent to the gold one is frequently a second valid answer, so labeling it negative pushes a relevant passage away. Second, in our development runs mined negatives reduced Recall@10 rather than improving it, consistent with the high terminological overlap of regulatory drafting. Equation~\eqref{eq:mnrl} already supplies $N-1$ negatives per query at no additional cost and led to more stable training.

\subsection{Stage 3: Hybrid Fusion and Context Assembly}
\label{subsec:hybrid_fusion}

Dense retrieval alone under-weights the exact identifiers (rule numbers, defined terms, entity names) that regulatory questions often hinge on. We therefore score each query with both retrievers, normalize each score by its per-query maximum,
\begin{equation}
\hat{s}_\bullet = \frac{s_\bullet}{\max(s_\bullet)+\epsilon}, \qquad \epsilon = 10^{-9},
\end{equation}
and fuse them linearly, as seen in \cite{wang2021interpolation}:
\begin{equation}
s_{\text{hybrid}} = \alpha\,\hat{s}_{\text{dense}} + (1-\alpha)\,\hat{s}_{\text{bm25}}. \label{eq:bm25fusion}
\end{equation}
We set $\alpha=0.8$, keeping the tuned dense retriever dominant while letting BM25 break ties on exact terminology.

The top $k=10$ passages are filtered before they reach the generator: passages scoring below 0.7 are dropped, and if consecutive ranked scores fall by more than 0.2 the remaining tail is truncated. At least one passage is always retained. This filtering is important because RePASs measures obligation coverage against the supplied context. Weakly relevant passages increase the number of obligations an answer is expected to cover and thus penalize the generator for retrieval errors.

\subsection{Generation with Prompting}

The first generator variant conditions a frozen instruction-tuned model on $(q, P_k)$. All models receive an identical system instruction, namely to act as a regulatory compliance assistant and synthesize every obligation in the supplied passages into one coherent answer, under three strategies: zero-shot; few-shot with three worked examples, one of which demonstrates the correct response to a question the context does not answer; and few-shot with chain-of-thought \cite{wei2023chainofthoughtpromptingelicitsreasoning,brown2020languagemodelsfewshotlearners,liu2021pretrainpromptpredictsystematic}, adding explicit reasoning instructions and constraints while still emitting only the final answer. Decoding is greedy with a 512-token budget.

\subsection{Generation with RAFT-LoRA}

The second variant adapts the generator. Following RAFT \cite{zhang2024raftadaptinglanguagemodel}, we build training instances from our own hybrid retriever so that the distractors the model learns to resist are the ones it will actually see. For a fraction $P=0.7$ of questions the context contains the oracle passage plus distractors; for the remaining $0.3$ it contains distractors only:
\begin{equation}
\begin{aligned}
70\%:\quad & Q_i + D_i^{*} + D_1 + D_2 + D_3 \rightarrow A_i^{*} \\
30\%:\quad & Q_i + D_1 + D_2 + D_3 + D_4 \rightarrow A_i^{*}
\end{aligned}
\end{equation}
Passage order is shuffled to suppress positional bias. Targets $A_i^{*}$ are produced by GPT-4o-mini \cite{openai2024gpt4ocard} as a teacher, prompted to emit a \textsc{Reason} section followed by an \textsc{Answer} section, to use only the supplied documents, and to state explicitly when the documents do not settle the question.

Two properties of this setup should be stated explicitly. The targets $A_i^{*}$ are teacher outputs from GPT-4o-mini rather than human-written gold answers, so what the adapted models learns is the teacher's answer behaviour, namely exhaustive enumeration of the obligations present in the supplied passages and explicit abstention when they are absent, over contexts that our own retriever produced. The RAFT training sets are also deliberately small, at 150 to 200 instances per corpus or under 1\% of the available training questions. We therefore present this as an efficient adaptation recipe, one teacher pass and a few hundred optimisation steps on a single GPU, and not as evidence that the resulting behaviour is independent of the teacher model or of how the training contexts were assembled.

Adaptation uses QLoRA \cite{dettmers2023qloraefficientfinetuningquantized}: the base model is frozen in 4-bit NF4, and rank-16 LoRA adapters are injected into the query and value projections. Loss is computed only over the reason and answer spans, with instruction, question, and context tokens masked, so the model is not trained to reproduce its input. Adapters are merged into the base weights for inference and also stored separately for reuse. Table~\ref{tab:lora} lists the configuration.

\begin{table}[t]
\centering
\caption{RAFT dataset construction and QLoRA adaptation settings, shared across all adapted models.}
\label{tab:lora}
\footnotesize
\setlength{\tabcolsep}{4pt}
\begin{tabular}{l l}
    \toprule
    Component & Setting \\
    \midrule
    Base models & Qwen2.5-7B, Teuken-7B v0.6, Gemma-2-2B \\
    \rowcolor{gray!10}Quantisation & 4-bit NF4, FP16 compute \\
    LoRA rank $r$ / $\alpha$ / dropout & 16 / 32 / 0.05 \\
    \rowcolor{gray!10}Target modules & \texttt{q\_proj}, \texttt{v\_proj} \\
    Optimiser & paged AdamW (8-bit) \\
    \rowcolor{gray!10}Learning rate / warm-up & $2\times10^{-4}$ / 0.05 \\
    Epochs / batch size & 3 / 1 \\
    \rowcolor{gray!10}Teacher model (CoT labels) & GPT-4o-mini \\
    Oracle-passage probability $P$ & 0.7 \\
    \rowcolor{gray!10}Loss & completion-only cross-entropy \\
    \bottomrule
\end{tabular}

\end{table}

\subsection{Evaluation}

Retrieval is measured by Recall@10, MRR@10, and nDCG@10 against the annotated gold passages.

Generation is measured with RePASs \cite{gokhan2024riragregulatoryinformationretrieval}, which scores an answer against the retrieved context rather than against a reference string. For answer sentences $a_i$ and passage sentences $p_j$, entailment and contradiction take the best-matching passage sentence per answer sentence,
\begin{equation}
E_s = \tfrac{1}{N}\sum_{i}\max_{j} P_{\text{ent}}(p_j, a_i), \quad
C_s = \tfrac{1}{N}\sum_{i}\max_{j} P_{\text{con}}(p_j, a_i),
\end{equation}
while obligation coverage first classifies which context sentences state obligations, using a LegalBERT obligation classifier, and then checks how many are entailed by some answer sentence above a 0.7 threshold:
\begin{equation}
OC_s = \tfrac{1}{M}\sum_{k}\mathbf{1}\!\left(\max_{l} P_{\text{ent}}(o_k, a_l) > 0.7\right).
\end{equation}
The composite is
\begin{equation}
\text{RePASs} = \frac{E_s - C_s + OC_s + 1}{3} \in [0,1].
\label{eq:repass}
\end{equation}
Entailment and contradiction come from \texttt{nli-deberta-v3-xsmall} and obligation matching from \texttt{deberta-large-mnli}. We report all three components separately because Equation~\eqref{eq:repass} combines terms that can move in opposite directions, so two systems with the same composite score may behave quite differently.

\section{Experiments}

\subsection{Setup}

All experiments run on single NVIDIA T4, L4, or A100 GPUs, with A100s reserved for generation and LoRA training. Generators are loaded in 4-bit NF4 quantization via \texttt{bitsandbytes}, so every model in Table~\ref{tab:obliqa-gen} fits on one commodity accelerator. We evaluate five compact instruction-tuned models: Qwen2.5-7B \cite{bai2023qwentechnicalreport}, Teuken-7B v0.6 \cite{ali2025teuken7bbaseteuken7binstructeuropean}, Gemma-2-2B \cite{gemmateam2024gemma2improvingopen}, Gemma-3-12B \cite{gemmateam2025gemma3technicalreport}, and DeepSeek-R1-Distill-Llama-8B \cite{Guo_2025}.

All reported numbers are computed on the validation split of each corpus; the two-document ObliQA test split is held out and untouched. Retrieval is evaluated exhaustively against the complete corpora, on 3,733 ObliQA question--passage pairs drawn from 3,191 questions and on 427 Australian queries. Generation is evaluated on a fixed sample of 150 validation questions per configuration, held constant across models and strategies; the sample is a concession to the cost of scoring long-form answers with two NLI models, and we return to it in Section~\ref{sec:limitations}.

\subsection{Retrieval}
\label{sec:retrieval-results}

\begin{figure*}[t]
\centering
\includegraphics[width=\textwidth]{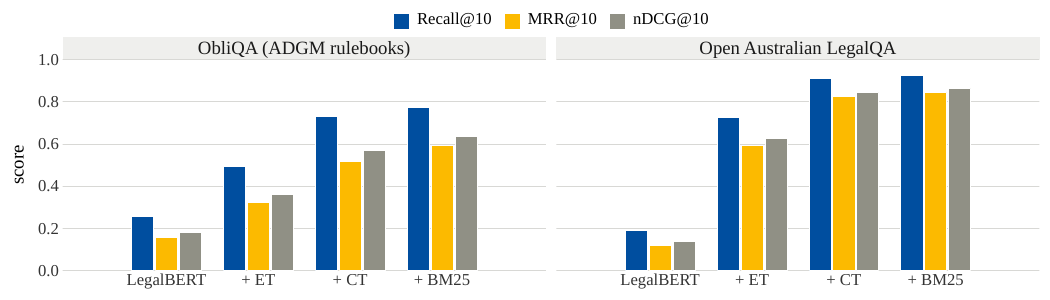}
\caption{Retrieval quality after each adaptation stage. The ordering of stages is identical on both corpora, but their relative contributions are not: entailment tuning is decisive on Australian case law, contrastive tuning on the ADGM rulebooks. Values are those of Table~\ref{tab:retriever}.}
\label{fig:retriever}
\end{figure*}

Table~\ref{tab:retriever} reports the staged ablation and Figure~\ref{fig:retriever} plots it.

\begin{table}[t]
\centering
\caption{Staged retriever ablation. Each stage is initialized from the previous one; the hybrid adds BM25 at $\alpha=0.8$.}
\label{tab:retriever}
\footnotesize
\setlength{\tabcolsep}{3.2pt}
\begin{tabular}{l c c c}
    \toprule
    Retriever stage & R@10 & MRR@10 & nDCG@10 \\
    \midrule
    \multicolumn{4}{l}{\textit{ObliQA (ADGM rulebooks)}} \\
    LegalBERT (baseline)      & 0.256 & 0.158 & 0.181 \\
    \rowcolor{gray!10}+ entailment tuning (ET)  & 0.495 & 0.322 & 0.363 \\
    + contrastive tuning (CT) & 0.732 & 0.518 & 0.569 \\
    \rowcolor{gray!10}+ BM25 fusion (hybrid)    & \textbf{0.774} & \textbf{0.594} & \textbf{0.638} \\
    \midrule
    \multicolumn{4}{l}{\textit{Open Australian LegalQA}} \\
    LegalBERT (baseline)      & 0.190 & 0.120 & 0.137 \\
    \rowcolor{gray!10}+ entailment tuning (ET)  & 0.726 & 0.595 & 0.626 \\
    + contrastive tuning (CT) & 0.911 & 0.824 & 0.845 \\
    \rowcolor{gray!10}+ BM25 fusion (hybrid)    & \textbf{0.923} & \textbf{0.846} & \textbf{0.865} \\
    \bottomrule
\end{tabular}

\end{table}

\say{Off-the-shelf} LegalBERT performs poorly as a retriever: it ranks the gold passage among the top ten for only one in four ObliQA queries and one in five Australian queries. Therefore, Domain-specific pretraining alone does not yield an embedding space suitable for retrieval.

Entailment tuning roughly doubles every ObliQA metric (Recall@10 $0.256 \rightarrow 0.495$) and has a far larger effect on Australian case law ($0.190 \rightarrow 0.726$). We attribute this difference to the question types. Australian questions name a case or instrument and ask what it provides, so a single passage decisively entails the answer and the premise--hypothesis objective aligns almost perfectly with the task. ADGM questions ask what a class of regulated entity must do, and several provisions bear on that, so entailment alone leaves ambiguity.

Contrastive tuning is the larger step on ObliQA ($0.495 \rightarrow 0.732$) and consolidates the Australian gain ($0.726 \rightarrow 0.911$). We attribute this to the model learning to distinguish between passages with similar regulatory vocabulary, which is the main difficulty in ObliQA. BM25 fusion adds a final $+0.042$ on ObliQA and $+0.012$ on Australian data, with a disproportionate effect on ranking quality: on ObliQA, Recall@10 rises 5.7\% relative while MRR@10 rises 14.7\%, i.e.\ BM25 mostly promotes passages the dense model had already retrieved but ranked too low. That is the expected signature of exact-term evidence, and it is worth more than the recall number suggests, because a generator reads the top of the list most attentively.

\begin{table}[t]
\centering
\caption{ObliQA retrieval against standard baselines, full validation split (3,733 question--passage pairs, 13,705-passage corpus).}
\label{tab:retriever-baselines}
\footnotesize
\setlength{\tabcolsep}{4pt}
\begin{tabular}{l l c c c}
    \toprule
    Retriever & Type & R@10 & MRR@10 & nDCG@10 \\
    \midrule

    BM25 & lexical & 0.678 & 0.504 & 0.546 \\
    \rowcolor{gray!10}BGE-M3 & dense & 0.711 & 0.544 & 0.584 \\
    BGE-base-en-v1.5 & dense & 0.719 & 0.549 & 0.590 \\
    \rowcolor{gray!10}E5-base-v2 & dense & 0.721 & 0.557 & 0.597 \\
    E5-large-v2 & dense & 0.758 & \textbf{0.595} & 0.635 \\
    \rowcolor{gray!10}Hybrid ET+CT (ours) & lexical+dense & \textbf{0.774} & 0.594 & \textbf{0.638} \\

    \bottomrule
\end{tabular}

\end{table}

Table~\ref{tab:retriever-baselines} places the hybrid retriever against standard alternatives on ObliQA. BM25 alone reaches 0.678, within 0.04 Recall@10 of BGE-M3 and BGE-base despite learning nothing about the domain, which underlines the importance of exact terminology in regulatory retrieval. Among dense baselines E5-large-v2 is strongest at 0.758. Our hybrid reaches 0.774 Recall@10 and 0.638 nDCG@10, above every baseline, while its dense component is a BERT-base encoder, roughly a third of E5-large's parameters, adapted on in-domain data alone. We note that E5-large-v2 attains a marginally higher MRR@10 (0.595 vs.\ 0.594); the two systems are equivalent on first-hit rank and differ in the depth of the ranking.

\subsection{Generation on ObliQA}

\begin{figure*}[t]
\centering
\includegraphics[width=\textwidth]{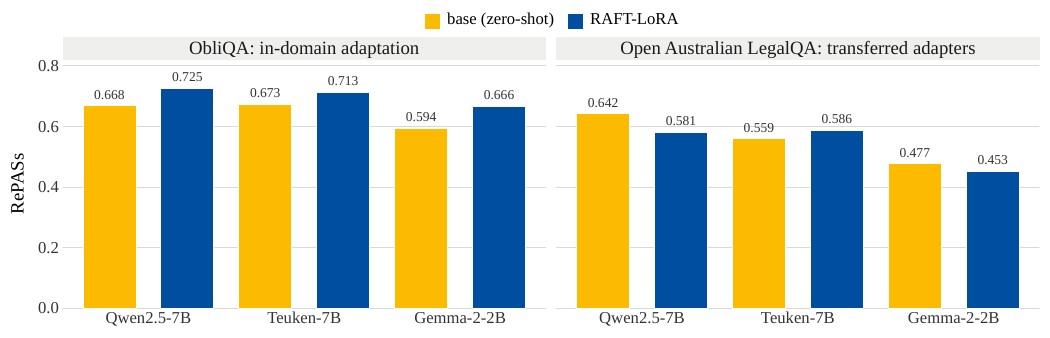}
\caption{RAFT-LoRA against the zero-shot base model. Left: adapters trained and evaluated on ObliQA improve every model. Right: the same ObliQA-trained adapters evaluated on Australian case law, where only Teuken retains a gain. Values are those of Tables~\ref{tab:obliqa-gen} and~\ref{tab:auslegal-gen}.}
\label{fig:raft}
\end{figure*}

Table~\ref{tab:obliqa-gen} reports RePASs and its components for every model and strategy; Figure~\ref{fig:raft} summarizes the base-versus-adapted comparison on both corpora. Our four key findings on ObliQA are:

\begin{table}[t]
\centering
\caption{Answer generation on ObliQA, 150 evaluation questions, hybrid retriever context. Best RePASs per model in bold.}
\label{tab:obliqa-gen}
\footnotesize
\setlength{\tabcolsep}{3.4pt}
\begin{tabular}{l l c c c c}
    \toprule
    Model & Strategy & $E_s\uparrow$ & $C_s\downarrow$ & $OC_s\uparrow$ & RePASs$\uparrow$ \\
    \midrule

    Qwen2.5-7B & zero-shot & 0.894 & 0.189 & 0.301 & 0.668 \\
    & few-shot & 0.971 & 0.167 & 0.371 & \textbf{0.725} \\
    & few-shot + CoT & 0.957 & 0.154 & 0.359 & 0.721 \\
    & RAFT-LoRA & 0.926 & 0.157 & 0.409 & \textbf{0.725} \\

    \rowcolor{gray!10}Teuken-7B & zero-shot & 0.944 & 0.256 & 0.332 & 0.673 \\
    \rowcolor{gray!10}v0.6 & few-shot & 0.928 & 0.200 & 0.259 & 0.662 \\
    \rowcolor{gray!10}& few-shot + CoT & 0.929 & 0.225 & 0.261 & 0.655 \\
    \rowcolor{gray!10}& RAFT-LoRA & 0.877 & \textbf{0.134} & 0.397 & \textbf{0.713} \\

    Gemma-2-2B & zero-shot & 0.935 & 0.317 & 0.167 & 0.594 \\
    & few-shot & 0.908 & 0.342 & 0.128 & 0.565 \\
    & few-shot + CoT & 0.900 & 0.326 & 0.113 & 0.562 \\
    & RAFT-LoRA & 0.934 & 0.212 & 0.277 & \textbf{0.666} \\

    \rowcolor{gray!10}Gemma-3-12B & zero-shot & \textbf{0.961} & 0.312 & 0.328 & 0.659 \\
    \rowcolor{gray!10}& few-shot & 0.930 & 0.264 & 0.333 & 0.666 \\
    \rowcolor{gray!10}& few-shot + CoT & 0.911 & 0.182 & 0.353 & \textbf{0.694} \\
    \rowcolor{gray!10}& RAFT-LoRA & \multicolumn{4}{c}{\textit{unstable adaptation -- excluded}} \\

    DeepSeek-R1 & zero-shot & 0.693 & 0.365 & \textbf{0.877} & \textbf{0.735} \\
    Distill-8B & few-shot & 0.605 & 0.335 & 0.836 & 0.702 \\
    & few-shot + CoT & 0.629 & 0.365 & 0.876 & 0.713 \\
    & RAFT-LoRA & \multicolumn{4}{c}{\textit{unstable adapter merge -- excluded}} \\

    \bottomrule
\end{tabular}

\end{table}

\textbf{1. Prompting does not generalize across models.} Qwen2.5 improves from 0.668 zero-shot to 0.725 with few-shot examples. Teuken degrades ($0.673 \rightarrow 0.662$), Gemma-2 degrades further ($0.594 \rightarrow 0.565$), and DeepSeek-R1 also loses ground ($0.735 \rightarrow 0.702$). Adding chain-of-thought on top of few-shot helps only Gemma-3 ($0.666 \rightarrow 0.694$) and DeepSeek-R1 ($0.702 \rightarrow 0.713$), and is neutral-to-negative elsewhere. The component columns explain why: for Teuken and Gemma-2 the examples cut obligation coverage sharply (Teuken $0.332 \rightarrow 0.259$, Gemma-2 $0.167 \rightarrow 0.128$) without a compensating drop in contradiction. Our three demonstrations are short, and models with weaker instruction-following imitate their length and shape rather than their reasoning, truncating the exhaustive obligation enumeration that RePASs rewards. This is consistent with evidence that demonstrations mostly convey format \cite{min2022rethinkingroledemonstrationsmakes}, and it means that few-shot prompting is not a safe default in this domain: it must be validated per model.

\textbf{2. RAFT-LoRA improves every model it can be applied to.} Qwen2.5 gains $+0.057$ (0.668 $\rightarrow$ 0.725), Teuken $+0.040$ (0.673 $\rightarrow$ 0.713), and Gemma-2 $+0.072$ (0.594 $\rightarrow$ 0.666). The mechanism is visible in the components and is the same in all three cases: obligation coverage rises substantially (Qwen $0.301\rightarrow0.409$, Teuken $0.332\rightarrow0.397$, Gemma-2 $0.167\rightarrow0.277$) while contradiction falls (Qwen $0.189\rightarrow0.157$, Teuken $0.256\rightarrow0.134$, Gemma-2 $0.317\rightarrow0.212$). This matches the goal of the RAFT objective, namely covering what the context supports without adding unsupported claims. Interestingly, Teuken's entailment score \emph{drops} from 0.944 to 0.877 at the same time. Teuken's base answers are short and closely paraphrase one passage, which scores well on entailment and poorly on coverage; the adapted model writes longer, multi-provision answers in which some sentences are aggregations rather than restatements. For compliance use the adapted behaviour is preferable, and the composite metric agrees, but the raw entailment column would have suggested a regression. 

Gemma-2-2B, the smallest model, benefits most from adaptation. In practice, this means that a few hundred training steps of retrieval-aware adaptation can noticeably improve grounding in a 2B model.

\textbf{3. Adaptation is not universally available.} Two of the five models could not be adapted. Merging LoRA adapters into DeepSeek-R1-Distill produced inconsistent generation at inference, and Gemma-3 showed near-zero loss movement across epochs, indicating that the target sequence was recoverable from the input and no useful gradient signal was present. We exclude both models from the comparison. For deployment, however, this shows that LoRA adaptation and adapter merging do not work equally reliably across architectures.

\textbf{4. The strongest single ObliQA number belongs to an unadapted model.} DeepSeek-R1-Distill reaches 0.735 zero-shot, above every RAFT-LoRA result. Its profile is unlike the others: obligation coverage 0.877, far higher than any other system, with entailment of only 0.693 and contradiction of 0.365. A reasoning-distilled model enumerates context obligations exhaustively, which RePASs rewards twice: directly through $OC_s$, and indirectly because long enumerations dilute the per-sentence contradiction average. Its lower entailment, meanwhile, shows that much of what it writes is not directly supported by any single passage. The composite score cannot tell whether these answers are more useful for compliance than Qwen's more conservative ones.

\subsection{Cross-Domain Transfer}

The Australian results in Table~\ref{tab:auslegal-gen} serve two purposes: they test whether the retriever and prompting conclusions hold on a different legal genre, and they test whether the ObliQA-trained adapters transfer. The adapters are reused unchanged; no Australian RAFT data was generated.

\begin{table}[t]
\centering
\caption{Answer generation on Open Australian LegalQA, 150 evaluation questions. $^{\dagger}$ marks adapters trained on ObliQA and applied without retraining.}
\label{tab:auslegal-gen}
\footnotesize
\setlength{\tabcolsep}{3.4pt}
\begin{tabular}{l l c c c c}
    \toprule
    Model & Strategy & $E_s\uparrow$ & $C_s\downarrow$ & $OC_s\uparrow$ & RePASs$\uparrow$ \\
    \midrule

    Qwen2.5-7B & zero-shot & 0.964 & 0.414 & 0.377 & \textbf{0.642} \\
    & few-shot & 0.964 & 0.402 & 0.350 & 0.637 \\
    & few-shot + CoT & 0.952 & 0.492 & 0.242 & 0.567 \\
    & RAFT-LoRA$^{\dagger}$ & 0.957 & 0.451 & 0.238 & 0.581 \\

    \rowcolor{gray!10}Teuken-7B & zero-shot & 0.943 & 0.552 & 0.288 & 0.559 \\
    \rowcolor{gray!10}v0.6 & few-shot & 0.941 & 0.578 & 0.212 & 0.525 \\
    \rowcolor{gray!10}& few-shot + CoT & 0.945 & 0.587 & 0.209 & 0.522 \\
    \rowcolor{gray!10}& RAFT-LoRA$^{\dagger}$ & 0.951 & 0.463 & 0.271 & \textbf{0.586} \\

    Gemma-2-2B & zero-shot & 0.922 & 0.639 & 0.151 & \textbf{0.477} \\
    & few-shot & 0.892 & 0.676 & 0.116 & 0.444 \\
    & few-shot + CoT & 0.881 & 0.676 & 0.104 & 0.436 \\
    & RAFT-LoRA$^{\dagger}$ & 0.940 & 0.592 & 0.013 & 0.453 \\

    \rowcolor{gray!10}Gemma-3-12B & zero-shot & \textbf{0.964} & 0.515 & 0.237 & \textbf{0.562} \\
    \rowcolor{gray!10}& few-shot & 0.959 & 0.545 & 0.200 & 0.538 \\
    \rowcolor{gray!10}& few-shot + CoT & 0.955 & 0.571 & 0.176 & 0.520 \\
    \rowcolor{gray!10}& RAFT-LoRA & \multicolumn{4}{c}{\textit{unstable adaptation -- excluded}} \\

    DeepSeek-R1 & zero-shot & 0.571 & 0.610 & 0.850 & 0.604 \\
    Distill-8B & few-shot & 0.552 & 0.563 & 0.844 & 0.611 \\
    & few-shot + CoT & 0.552 & 0.573 & \textbf{0.862} & \textbf{0.613} \\
    & RAFT-LoRA & \multicolumn{4}{c}{\textit{unstable adapter merge -- excluded}} \\

    \bottomrule
\end{tabular}

\end{table}

The findings on prompting are even clearer here: zero-shot is the best prompted strategy for Qwen, Teuken, Gemma-2, and Gemma-3, and few-shot and chain-of-thought degrade every one of them. Only DeepSeek-R1 prefers chain-of-thought (0.613), again through obligation coverage.

The adapters, however, mostly do not transfer. Qwen drops from 0.642 to 0.581 and Gemma-2 from 0.477 to 0.453; only Teuken improves, 0.559 to 0.586. The component scores show why: Gemma-2's obligation coverage collapses to 0.013, because the adapted model reproduces the terse rulebook-style answer it was trained to produce, which covers almost nothing in a long case-law passage. Qwen's coverage falls from 0.377 to 0.238 for the same reason. Teuken is the exception because its adaptation gain came predominantly from contradiction reduction ($0.552 \rightarrow 0.463$), and learning not to assert unsupported claims is a genre-independent behaviour, whereas learning the shape of an ADGM obligation answer is not.

We attribute this to two structural differences between the corpora. ObliQA passages are short numbered provisions with an explicit normative operator, and the mapping from retrieved passage to answer is close to deterministic. Australian passages are extended excerpts of judicial reasoning in which the answer is distributed across the passage and rarely stated as an obligation at all. Contradiction scores are also consistently higher on the Australian data: every model contradicts its context far more often on Australian data (0.40--0.68) than on ObliQA (0.13--0.37), even though retrieval is much better there (Recall@10 0.923 vs.\ 0.774). Long argumentative passages contain positions the court ultimately rejects, and an NLI model scoring a summary against every sentence of such a passage will find contradictions that are not errors. This is a limitation of sentence-level NLI metrics on case law, and it means Table~\ref{tab:auslegal-gen} should be read within-column rather than against Table~\ref{tab:obliqa-gen}.

The practical implication is that RAFT-LoRA adapters are corpus-specific artifacts. A firm deploying this pipeline across several regulatory regimes should expect to build RAFT data per regime rather than to reuse one adapter.

\subsection{Is Retrieval Doing the Work?}
\label{sec:closedbook}

To check that the retrieval pipeline is responsible for the answer quality we measure, we ran Qwen2.5 and Teuken closed-book: same prompt, same questions, no passages. RePASs is still computed against the retrieved passages the model never saw. These runs cover 132 and 89 validation questions respectively rather than the full 150, so they are indicative rather than precise; the effect below is far larger than that sample difference can explain.

\begin{table}[t]
\centering
\caption{Closed-book control on ObliQA. Scores are computed against retrieved passages that the closed-book models did not receive. Closed-book runs cover 132 (Qwen) and 89 (Teuken) validation questions.}
\label{tab:closedbook}
\footnotesize
\setlength{\tabcolsep}{4pt}
\begin{tabular}{l l c c c c}
    \toprule
    Model & Setting & $E_s\uparrow$ & $C_s\downarrow$ & $OC_s\uparrow$ & RePASs$\uparrow$ \\
    \midrule

    Qwen2.5-7B & closed-book & 0.883 & \textbf{0.081} & 0.169 & 0.657 \\
    \rowcolor{gray!10}& RAG (zero-shot) & 0.894 & 0.189 & \textbf{0.301} & \textbf{0.668} \\

    Teuken-7B v0.6 & closed-book & 0.836 & \textbf{0.096} & 0.247 & 0.662 \\
    \rowcolor{gray!10}& RAG (zero-shot) & 0.944 & 0.256 & \textbf{0.332} & \textbf{0.673} \\

    \bottomrule
\end{tabular}

\end{table}

Table~\ref{tab:closedbook} shows the results. Closed-book Qwen scores 0.657 compared to 0.668 with retrieval, and closed-book Teuken 0.662 compared to 0.673. A difference of about 0.01 clearly understates the gap between an answer grounded in the ADGM rulebook and one generated from parametric knowledge alone. Inspection of the closed-book outputs confirms that they are plausible-sounding regulatory prose that misstates obligations, omits binding conditions, and cites nothing.

The components show how the score is obtained. Closed-book contradiction is very low (0.081 and 0.096, against 0.189 and 0.256 with retrieval), because an answer that never commits to a specific provision has little to contradict; and $E_s$ takes a maximum over passage sentences, so generic regulatory statements find some sentence that entails them. Obligation coverage does fall as expected (Qwen $0.301 \rightarrow 0.169$), but it is one of three terms in Equation~\eqref{eq:repass} and the contradiction reduction offsets most of it.

We draw two conclusions from this. First, RePASs should not be used on its own to rank regulatory QA systems, and results should be reported together with a closed-book baseline. Second, the main benefit of retrieval in our setting is attributability, i.e., the ability to trace an answer back to a specific provision. Compliance applications require this property, but RePASs does not capture it. Of its three components, only obligation coverage clearly separates the retrieval and closed-book settings.

\subsection{Limitations}
\label{sec:limitations}

Several constraints bound these results. Generation is scored on 150 questions per configuration, and the closed-book control on fewer still; differences of a few thousandths of RePASs, such as Qwen's tie between few-shot and RAFT-LoRA, are within noise, and we make no claims at that resolution. The leakage-free re-split discards 989 multi-passage question--passage pairs, tilting ObliQA toward single-passage questions and away from the multi-provision reasoning that real compliance work demands. RePASs, as Section~\ref{sec:closedbook} shows, is not a sufficient measure of grounding; an attribution-sensitive metric or an LLM-as-judge protocol \cite{li-etal-2025-generation} would be a better instrument, and human expert evaluation better still. Finally, the two corpora studied here are both English and both drawn from public sources; internal policy documents, tabular disclosures, and multilingual rulebooks remain untested.

\section{Conclusion}

We investigated how much of the gap between compact language models and reliable regulatory question answering can be closed by adapting the pipeline instead of scaling the model. The retrieval half of that gap closes substantially and is directly measured: a LegalBERT encoder that ranks the correct Abu Dhabi Global Market (ADGM) provision among its top ten for one query in four does so for three queries in four after entailment tuning, contrastive tuning, and BM25 fusion, ahead of E5-large-v2 at a fraction of the parameters. The generation half improves on the metric available to us, in that retrieval-aware LoRA adaptation raises composite RePASs for every compact generator we could adapt, most for the smallest, by increasing obligation coverage and reducing contradictions on a few hundred teacher-labelled examples. We describe this as improved in-domain answer behaviour rather than as improved grounding, because our closed-book control shows that RePASs does not distinguish the two.

Three conclusions are supported by these experiments. First, domain-adapted retrieval is effective: staged adaptation of a small in-domain encoder outperforms both a strong lexical baseline and larger general-purpose dense encoders on regulatory text. Second, RAFT-LoRA improves compact generators in-domain, and in-domain only: adapters trained on ADGM rulebook data lost most of their benefit on Australian case law, so deployments spanning several regulatory regimes should plan for regime-specific RAFT data rather than adapter reuse. Third, current evaluation metrics can substantially understate the value of retrieval. A closed-book model that received no passages, cited nothing, and misstated obligations scored within 0.011 RePASs of the full pipeline, so RePASs cannot on its own validate a compliance system. Therefore, Retrieval remains essential for transparency and traceability, but regulatory NLP needs evaluation protocols that can measure these properties.

\section*{Acknowledgments}

This research has been partially funded by the Federal Ministry of Education and Research of Germany and the state of North-Rhine Westphalia as part of the Lamarr-Institute for Machine Learning and Artificial Intelligence.

For this paper, Anthropic Claude Opus 5 \cite{anthropic2026opus5} was employed to assist in refining and improving the text throughout all sections of this paper. The authors retain full responsibility for the accuracy, integrity, and originality of the work.

\bibliographystyle{unsrtnat}
\bibliography{bib}

\end{document}